\documentclass[10pt,twocolumn,letterpaper]{article}

\usepackage{wacv}              

\usepackage{graphicx}
\usepackage{amsmath}
\usepackage{amssymb}
\usepackage{booktabs}

\usepackage[pagebackref,breaklinks,colorlinks]{hyperref}

\usepackage[capitalize]{cleveref}
\crefname{section}{Sec.}{Secs.}
\Crefname{section}{Section}{Sections}
\Crefname{table}{Table}{Tables}
\crefname{table}{Tab.}{Tabs.}

\usepackage{url}
\usepackage{amssymb}
\usepackage[utf8]{inputenc} 
\usepackage[T1]{fontenc}    
\usepackage{url}            
\usepackage{booktabs}       
\usepackage{amsfonts}       
\usepackage{nicefrac}       
\usepackage{microtype}      
\usepackage{bm}
\usepackage{enumitem}
\usepackage{graphicx}
\usepackage{amsmath}

\usepackage{array}
\usepackage{multirow}
\usepackage{wrapfig,lipsum,booktabs}
\usepackage{caption}
\usepackage{algorithm}
\usepackage{algorithmic}
\usepackage[dvipsnames]{xcolor}

\newtheorem{theorem}{Theorem}
 
\newtheorem{proposition}[theorem]{Proposition}

\newtheorem{problem}[theorem]{Problem}

\usepackage{enumitem}

\def\wacvPaperID{778} 
\def\confName{WACV}
\def\confYear{2024}

\begin{document}

\title{DECDM: Document Enhancement using Cycle-Consistent Diffusion Models}

\author{Jiaxin Zhang \quad 
Joy Rimchala \quad
Lalla Mouatadid \quad 
Kamalika Das \quad
Sricharan Kumar \\
Intuit AI Research \\
{\tt\small \{jiaxin\_zhang, Joy\_Rimchala, Lalla\_Mouatadid, kamalika\_das, sricharan\_kumar\}@intuit.com}
}

\maketitle
\begin{abstract} 
The performance of optical character recognition (OCR) heavily relies on document image quality, which is crucial for automatic document processing and document intelligence. However, most existing document enhancement methods require supervised data pairs, which raises concerns about data separation and privacy protection, and makes it challenging to adapt these methods to new domain pairs. To address these issues, we propose DECDM, an end-to-end document-level image translation method inspired by recent advances in diffusion models. Our method overcomes the limitations of paired training by independently training the source (noisy input) and target (clean output) models, making it possible to apply domain-specific diffusion models to other pairs. DECDM trains on one dataset at a time, eliminating the need to scan both datasets concurrently, and effectively preserving data privacy from the source or target domain. We also introduce simple data augmentation strategies to improve character-glyph conservation during translation. We compare DECDM with state-of-the-art methods on multiple synthetic data and benchmark datasets, such as document denoising and {\color{black}shadow} removal, and demonstrate the superiority of performance quantitatively and qualitatively.  
\end{abstract}
\section{Introduction}
\label{sec:intro}
\begin{figure}[h!]
    \centering
    \includegraphics[width=0.49\textwidth]{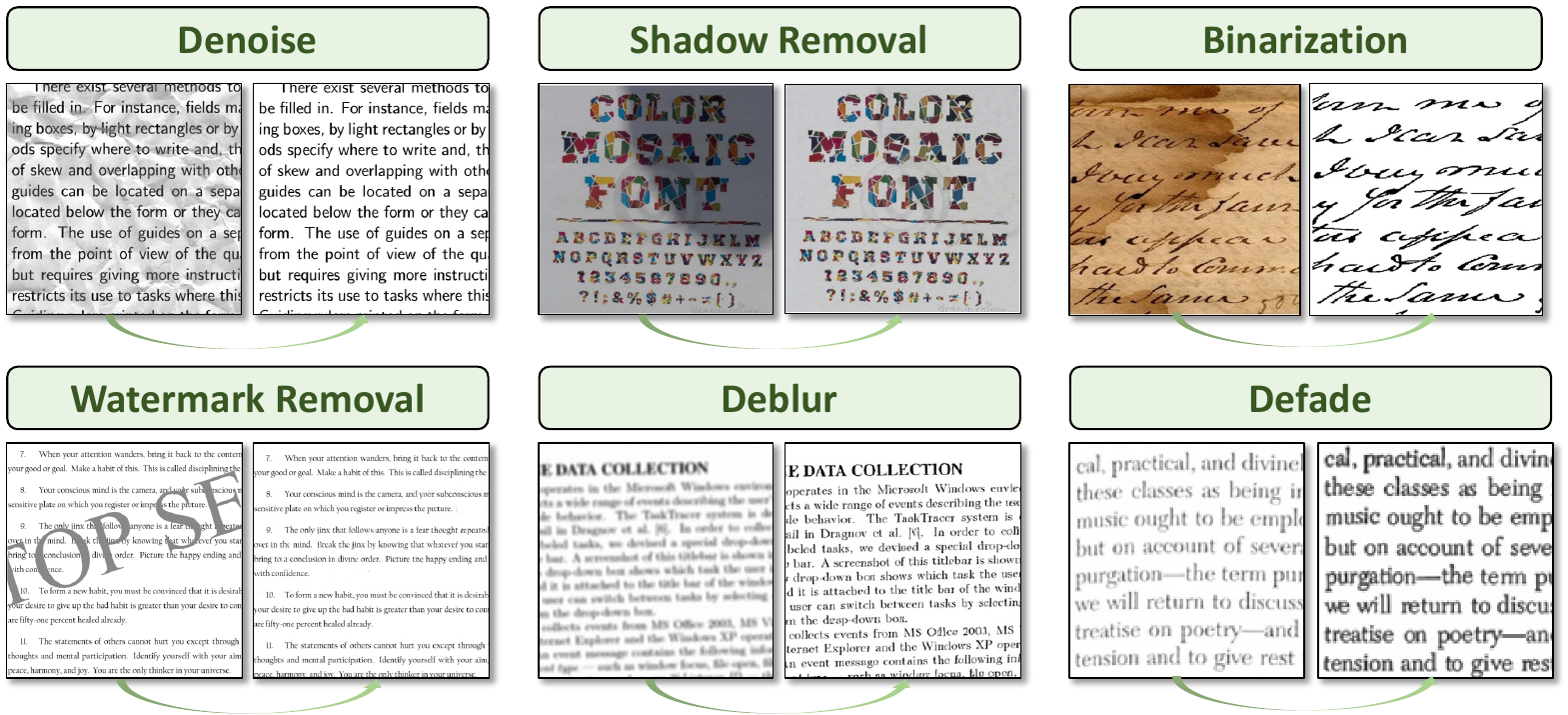}
    \caption{A performance overview of our DECDM methods on document enhancement tasks, including denoising, shadow removal, binarization, watermark removal, deblur, and defade. }
    \label{fig:overview1}
\end{figure}

In our daily lives, we encounter a large number of documents, such as receipts, invoices, and tax forms, that are often degraded in various ways, including noise, blurring, fading, watermarks, shadows, and more, as shown in Fig.~\ref{fig:overview1}. These degradations can make the documents difficult to read and can significantly impair the performance of OCR systems. Automatic document processing is the first step in document intelligence and aims to enhance document quality using advanced image processing techniques such as denoising, restoration, and deblurring. However, applying these techniques directly to document enhancement may not be effective due to the unique challenges posed by text documents. Unlike typical image restoration tasks, where the degradation function is known and the recovery of the image task can be translated into solving an inverse problem such as inpainting, deblurring/super-resolution, and colorization, real-world document enhancement is a blind denoising process with an unknown degradation function, making it even more challenging. Many state-of-the-art methods have been proposed that rely on assumptions and prior information \cite{song2021solving, kawar2022denoising}, but there is still a need for more effective techniques that can handle unknown degradation functions.
\begin{figure*}[h!]
    \centering
    \includegraphics[width=0.98\textwidth]{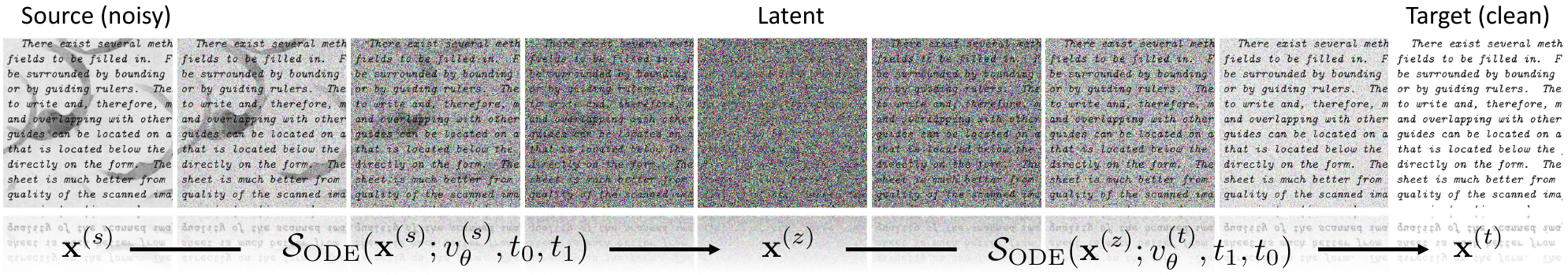}
    \caption{Cycle-Consistent Diffusion Models leverages two deterministic diffusions through ODEs for unpaired document-level image-to-image translation. Given source data $\mathbf x^{(s)}$, the source diffusion model $v_{\theta}^{(s)}$ runs in the forward direction to convert it to the latent space $\mathbf x^{(z)}$, while the target diffusion model $v_{\theta}^{(t)}$ reverse ODE to construct the target document-level images $\mathbf x^{(t)}$. $t_0$ and $t_1$ are the starting point and ending point, typically setting to $t_0 = 0 $ and $t_1 = 1$. }
    \label{fig:DECDM}
\end{figure*}

Deep learning has led to the development of discriminative models based on convolutional neural networks (CNNs) \cite{zhang2017beyond} and auto-encoder (AE) architectures \cite{xie2012image}, which are important for solving image restorations. However, these methods require noisy/clean paired image data, which is difficult to obtain in real-world applications. Existing benchmark datasets \cite{anvari2021survey} collect clean documents and add synthetic noise, but these do not always accurately represent real-world noise or degradation. To address this, recent works have proposed unpaired ideas based on generative models, such as generative adversarial networks (GANs) \cite{goodfellow2020generative}, which transfer images from one domain to another while preserving content representation \cite{zhu2017unpaired}. Document denoising can be achieved by transferring from a noisy style to a clean style while preserving the text content. However, these models typically require minimizing an adversarial loss between a specific pair of source and target datasets \cite{sharma2018learning}, which has limitations in training instability and potential data privacy leakage \cite{su2022dual}.

\begin{table*}[!h]
\resizebox{\linewidth}{!}{
\centering
\begin{tabular}{@{}c|c|ccc|cccccc@{}}
\toprule
\multirow{2}{*}{Methods} & \multirow{2}{*}{\begin{tabular}[c]{@{}c@{}}Unpaired \\ or paired\end{tabular}} & \multicolumn{3}{c|}{Backbone Models} & \multicolumn{6}{c}{Document Enhancement Tasks}                                                                                                                      \\ \cmidrule(l){3-11} 
                         &                                                                                & GANs      & CNNs     & Transformers     & Denoise & \begin{tabular}[c]{@{}c@{}}Shadow \\ Removal\end{tabular} & Binarization & \begin{tabular}[c]{@{}c@{}}Watermark \\ Removal\end{tabular} & Deblur & Defade \\ \midrule
SCGAN \cite{xu2017learning} (ICCV 17') & Paired & \checkmark & - & - & - & - & - & - & \checkmark & - \\
SCDCA \cite{zhao2018skip} (ICPR 18') & Paired & - & \checkmark & - & \checkmark & - & - & - & \checkmark & -      \\ 
BEDSR-Net \cite{lin2020bedsr} (CVPR 20') & Paired & \checkmark & - & - & - & \checkmark & - & - & - & - \\
DE-GAN \cite{souibgui2020gan} (TPAMI 20') & Paired & \checkmark & - & - & - & - & - & \checkmark & \checkmark & -  \\
RED-Net \cite{calvo2019selectional} (PR 19') & Paired & - & \checkmark & - & - & - & \checkmark & - & - & - \\
SauvolaNet \cite{li2021sauvolanet} (ICDAR 21') & Paired & - & \checkmark & - & - & - & \checkmark & - & - & - \\
CharFormer \cite{shi2022charformer} (ACM MM 22') & Paired & - & - & \checkmark & \checkmark & - & - & - & - & - \\ 
DocEnTr \cite{souibgui2022docentr} (ICPR '22) & Paired & - & - & \checkmark & - & - & \checkmark & - & \checkmark & \checkmark \\
CycleGAN \cite{sharma2018learning} (ACCV 18') & Unpaired & \checkmark & - & - & - & - & - & \checkmark & \checkmark & \checkmark \\                          
CycleGAN-MOE \cite{gangeh2021end} (ICCV 21') & Unpaired & \checkmark & - & - & \checkmark & - & - & \checkmark & \checkmark & \checkmark      \\
\bottomrule
\end{tabular}}
\vspace{-2mm}
\caption{A summary of document enhancement methods, including unpaired/paired supervision, backbone models (CNNs, GANs, Transformers), and enhancement tasks (denoise, shadow removal, binarization, watermark removal, deblur, defade).}
\vspace{-3mm}
\label{tab:summary}
\end{table*}

Beyond both disadvantages of existing methods, the task of document enhancement presents several unique challenges compared to typical image translation problems. These include (1) {\em High-resolution}, which poses scalability challenges, leading to performance degradation and significant increases in training costs. (2) {\em Lack of large benchmark datasets}, which makes it infeasible to use large pre-trained models. While the success of large generative models such as Stable diffusion \cite{rombach2022high}, Dall·E \cite{ramesh2022hierarchical}, and Imagen \cite{saharia2022photorealistic} is largely attributed to large datasets, such as LAION-5B \cite{schuhmann2022laion}, there is currently no large pre-trained model available for document-level tasks. (3) {\em Character feature damage}. Unlike image translation at the pixel level, document-level image translation requires preserving original content such as characters and words while accounting for style differences in the background, i.e., noise to clean. Current methods only focus on pixel-level information and do not consider critical character features such as glyphs, resulting in character-glyph damage during the translation process \cite{shi2022charformer}.

In this work, we present DECDM, an unsupervised end-to-end document-level image translation method that addresses the challenges faced by existing document enhancement methods. Inspired by recent advances in diffusion models \cite{song2020denoising,song2021solving,su2022dual,wu2022unifying}, our approach independently trains the source (noisy) and target (clean) models, decoupling paired training and enabling the domain-specific diffusion models to remain applicable to other pairs. Specifically, we build DECDM based on denoising diffusion implicit models (DDIMs) \cite{song2020denoising}, which create a deterministic and reversible mapping between images and their latent representations, solved using ordinary differential equation (ODE) that forms the cornerstone. Translation with DECDM on a source-target pair requires two different ODEs: the source ODE encodes input images to the latent space, while the target ODE decodes images in the target domain, as shown in Fig.~\ref{fig:DECDM}.

Since training diffusion models are specific to individual domains and rely on no domain pair information, DECDM makes it possible to save a trained model of a certain domain for future use, when it arises as the source or target in a new pair. Pairwise translation with DECDM requires only a linear number of diffusion models, which can be further reduced with conditional models \cite{dhariwal2021diffusion}.  Additionally, the training process focuses on one dataset at a time and does not require scanning both datasets concurrently, preserving the data privacy of the source or target domain.

To overcome the challenges in document-level translation, we propose a simple data augmentation scheme to downscale the resolution of training data, while significantly increasing the dataset size. This approach reduces the diffusion training cost and improves the performance in learning character distribution benefiting from large datasets. Experimentally, we demonstrate the effectiveness of DECDM on a variety of document enhancement tasks, such as document denoising and document shadow removal, with qualitative and quantitative results that establish DECDM as a scalable, efficient, and reliable solution to the family of document enhancement approaches. {\color{black} DECEM is also well-suited for few-shot scenarios by leveraging unpaired training and sample efficiency in cycle-consistent diffusion models and data augmentation strategies. Beyond the denoising and removal tasks shown here, our proposed DECDM method can apply to broader few-shot document enhancement tasks in Fig.~\ref{fig:overview1}}.

\section{DECDM Method}
Our goal is to develop a cycle-consistent diffusion model for document enhancement by solving the following three core problems: (1) unpaired supervision, (2) enforcing cycle consistency, and (3) data privacy protection. Then we introduce the data augmentation strategies for dealing with the challenges of document datasets while improving character and word feature preservation. 

\subsection{Problem Formulation}
We first define the unpaired document enhancement task from a mathematical perspective as follows:
\begin{problem}
  \textnormal{(Unpaired Document Enhancement)}. Given two unpaired sets of documents, one set consisting of degraded documents $\mathcal X$ (source domain), and the other a collection of clean documents $\mathcal Y$ (target domain), our goal is to learn a mapping $\mathcal F: \mathcal X \rightarrow \mathcal Y$ such that the output $ \hat{y} = \mathcal{F}(\mathbf x), \mathbf x \in \mathcal X$, is indistinguishable from documents $\mathbf y \in \mathcal Y$ to classify $\hat{\mathbf y}$ apart from $\mathbf y$.
\end{problem}
The degraded documents include multiple types, e.g., noise, blurring, watermark, etc, as shown in Fig.~\ref{fig:overview1}. The mapping $\mathcal F$ should satisfy two conditions: content preservation and style transfer. The content refers to the character, text, numbers, tables, and figures in documents and the style transfer means the translation from degraded documents (source domain $\mathcal{X}$) to clean documents (target domain $\mathcal{Y}$). Our objective is therefore to convert the degraded documents in $\mathcal{X}$ while preserving their core contents in $\mathcal{Y}$. From the computer vision perspective, enhancement tasks can be essentially interpreted as document-level image-to-image translation. 

\begin{problem}
 \textnormal{(Cycle Consistency)}. Assuming we have a mapping $\mathcal F: \mathcal X \rightarrow Y$ and another mapping $\mathcal H: \mathcal Y \rightarrow \mathcal X$, then $\mathcal F$ and $\mathcal H$ should be inverse of each other, and both mappings should be bijective, i.e., satisfying 
 \begin{equation}
     \mathcal F(\mathcal H (\mathbf x)) \approx \mathbf x, \quad \mathcal H (\mathcal F (\mathbf y)) \approx \mathbf y \label{eq:cycle}
 \end{equation} 
\end{problem}
A desirable feature of image translation algorithms is the {\em cycle consistency} property \cite{zhu2017unpaired}, which transforms a sample in the source domain to the target domain, and then back to the source, will recover the original sample in the source domain. This property is critical to the adaptability guarantee, which empowers the domain-specific diffusion models to stay applicable in other pairs. 
A rigorous formulation is defined in Eq.~\eqref{eq:cycle}.
\begin{problem}
 \textnormal{(Data Privacy)}. In the training and translation process, source model $v_{\theta}^{(s)}$ and target model $v_{\theta}^{(t)}$ are decoupled and trained independently, while both source datasets $ \mathbf x\in \mathcal X$  and target datasets $\mathbf y \in \mathcal Y$ are private to each other. 
\end{problem}
Most image-to-image translation approaches strongly rely on joint training over data from both source domains and target domains. This leads to a significant challenge in preserving the privacy of domain data in a federated setting. An ideal method is to train the models independently on separate domain datasets such that data privacy is protected. 

\begin{figure*}[h!]
    \centering
    \includegraphics[width=0.98\textwidth]{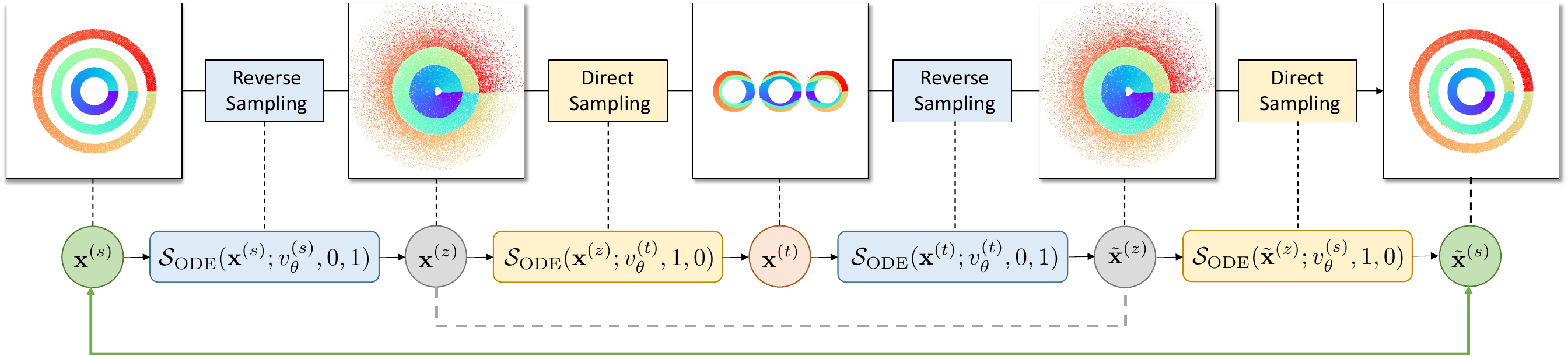}
    \caption{Cycle consistency illustration. Translation from the source domain (CR) to the target domain (PR) and then back to the source domain (CR) via the cycle-consistent diffusion models with reverse and direct sampling.}
    \label{fig:cycle}
\end{figure*}

\subsection{Cycle-Consistent Diffusion Models}
Diffusion Models \cite{sohl2015deep,ho2020denoising, song2019generative} aim at modeling a distribution $p_{\theta}(\mathbf x_0)$ to approximate the data distribution $q(\mathbf x_0)$ through diffusion and reversed generative processes. Song et al. \cite{song2020score} proposed a unified framework by leveraging Stochastic Differential Equations (SDEs) representation, which uses a forward and backward SDE to mathematically describe general diffusion processes: 
\begin{equation}
    \textnormal d \mathbf x = \mathbf f (\mathbf x, t) ~ \textnormal d t + g(t) ~ \textnormal d \mathbf w \label{eq:sde1}
\end{equation}
and reversed generative processes: 
\begin{equation}
    \textnormal d \mathbf x = [\mathbf f - g^2 \nabla_{\mathbf x} \log p_t(\mathbf x) ] ~\textnormal d t + g(t) ~ \textnormal d\mathbf w \label{eq:sde2}
\end{equation}
where $\mathbf f (\mathbf x, t)$ is the vector-valued coefficient, $\mathbf w$ is the standard Wiener process, $g(t)$ is the diffusion coefficient, and $\nabla_{\mathbf x} \log p_t(\mathbf x) $ is the score function of the noise perturbed data distribution.  Any diffusion process can be represented by a deterministic ODE \cite{song2020score}, named the probability flow (PF) ODE \cite{song2020score}, which enables uniquely identifiable encodings of data, and has the following form: 
\begin{equation}
     \textnormal d \mathbf x = \left[\mathbf f (\mathbf x, t) - \frac{1}{2} g(t)^2 \nabla_{\mathbf x} \log p_t(\mathbf x) \right] ~ \textnormal d t
\end{equation}
which is equivalent to the forward SDE in Eq.~\eqref{eq:sde1}. For conciseness, we use $\theta$-parameterized score networks $\mathbf s_{t, \theta} \approx \nabla_{\mathbf x} \log p_t(\mathbf x)$ to approximate the score function and use $v_{\theta} = \textup{d} \mathbf x / \textup{d}t$ to denote the $\theta$-parameterized model and use the symbol $\mathcal{S}_{\textup{ODE}}$ to denote the mapping from $\mathbf x^{(t_0)}$ to $\mathbf x^{(t_1)}$ and implement ODE solver in DDIMs \cite{song2020denoising}. 
\begin{equation}
\begin{aligned}
     \mathbf x(t_1) &= \mathcal{S}_{\textnormal{ODE}} (\mathbf x (t_0); v_{\theta}, t_0, t_1) 
     \\ &= \mathbf x (t_0) + \int_{t_0}^{t_1} v_{\theta}(t, \mathbf x(t)) ~ \textnormal d t
\end{aligned}
\end{equation}

In this work, we implement an ODE solver in DDIMs \cite{song2020denoising} where the generative sampling process is defined in a deterministic non-Markovian manner, which can be used for the reverse direction, deterministically noising an image to obtain the initial noise vector. This property is central to DECDM as we solve these ODEs for forward and reverse conversion between data and their latents. More details are provided in Appendix \ref{sec:ode_solver}. 

\paragraph{Cycle-Consistent Diffusion Models.}  
DECDM leverages the cycle-consistent diffusion models to perform unpaired document-level image translation, with two diffusion models trained independently on two separate domains. DECDM consists of two core steps, training, and translation, described in Algorithms 1 and 2. For training, DECDM first collects noisy data from the source domain $\mathbf x^{(s)} \sim p_{s}(\mathbf x)$, and clean data from the target domain $\mathbf x^{(t)} \sim p_{t}(\mathbf x)$, then train two diffusion models separately on the two domains and save them as  $v_{\theta}^{(s)}$ and $v_{\theta}^{(t)}$.  For translation, DECDM first runs $\mathcal{S}_{\textup{ODE}}$ in the source domain to obtain the latent encoding $\mathbf x^{(z)}$ of the image $\mathbf x^{(s)}$ at the end time $t_1$ via $\mathcal{S}_{\textnormal{ODE}} (\mathbf x^{(s)}; v_{\theta}^{(s)}, t_0, t_1)$. Then DECDM feds the source latent encoding $\mathbf x^{(z)}$ to $\mathcal{S}_{\textup{ODE}}$ with the target model $v_{\theta}^{(t)}$ to reconstruct the target image $\mathbf x^{(t)}$ via $\mathcal{S}_{\textnormal{ODE}} (\mathbf x^{(z)}; v_{\theta}^{(t)}, t_1, t_0)$, as illustrated in Fig.~\ref{fig:DECDM}. 

One of the important advantages of DECDM is the exact cycle consistency: transforms a sample in the domain $\mathcal{S}$ to the domain $\mathcal{T}$, and then back to $\mathcal{S}$,
will recover the original sample in $\mathcal{S}$. As probability flow ODEs are used, the cycle consistency property is guaranteed \cite{song2020score}. The following proposition validates the cycle consistency of DECDM. 

\begin{proposition} 
\textnormal{(Exact Cycle Consistency)}. Given a specific sample $\mathbf x^{(s)}$ from source domain $\mathcal X$, with a trained source model $v_{\theta}^{(s)}$ and a target model $v_{\theta}^{(s)}$, we define the forward cycle consistency 
\begin{equation}
\begin{aligned}
    \mathbf x^{(z)} = \mathcal{S}_{\textnormal{ODE}} (\mathbf x^{(s)}; v_{\theta}^{(s)}, t_0, t_1); \\
    \mathbf x^{(t)} = \mathcal{S}_{\textnormal{ODE}} (\mathbf x^{(z)}; v_{\theta}^{(t)}, t_1, t_0);
\end{aligned}
\end{equation}
and backward cycle consistency 
\begin{equation}
\begin{aligned}
    \tilde{\mathbf x}^{(z)} = \mathcal{S}_{\textnormal{ODE}} (\mathbf x^{(t)}; v_{\theta}^{(t)}, t_0, t_1); \\
    \tilde{\mathbf x}^{(s)} = \mathcal{S}_{\textnormal{ODE}} (\tilde{\mathbf x}^{(z)}; v_{\theta}^{(s)}, t_1, t_0);
\end{aligned}
\end{equation}
Assume zero discretization error, then we have $\mathbf x^{(s)} = \tilde{\mathbf x}^{(s)}$. 
\end{proposition}
In practice, we implement the ODE solver $\mathcal{S}_{\textnormal{ODE}}$ with DDIMs \cite{song2020denoising} which has reasonably small discretization errors. Thus DECDM incurs almost negligible cycle inconsistency.  Appendix \ref{sec:training} provides more details of training objectives, as illustrated in Algorithm \ref{algo:1}. Diffusion model training in DECDM returns the trained source model and target model, which are then employed for unpaired image translation as shown in Algorithm \ref{algo:2}.

\subsection{Data Privacy Protection}
The DECDM training process does not depend on knowledge of the domain pair a priori, while only source and target data are required. Both source and target diffusion models are trained independently. The DECDM translation process can be performed in a privacy-sensitive manner. For example, user A is the owner of the source domain and user B is the owner of the target domain. User A intends to translate the source images to the target domain in a private manner without releasing the source dataset. User B also wishes to make the target dataset private. In such a case, user  Acan can simply train a diffusion model with the source data, encode the data to the latent space, and only transmit the latent codes to user B. Then user B can use the pretrained diffusion models (using the target data) to convert the received latent code to a target image and send back to user A. The process only requires shared latent code from user A and a pretrained model from user B, which can be finished in a private platform, and both source and target datasets are private to the two parties. This is a significant advantage of DECDM over alternate methods, as we enable strong privacy protection of the datasets.  {\color{black} More discussions can be found in Appendix \ref{sec:privacy}}.

\subsection{Data Augmentation}
\begin{figure}[h!]
    \centering
    \includegraphics[width=0.49\textwidth]{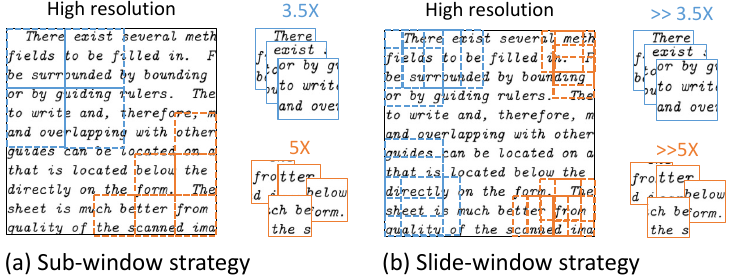}
    \caption{Data argumentation for document-level high-resolution images: (a) sub-window strategy and (b) slide-window strategy. }
    \label{fig:DA}
\end{figure}

Many document benchmark datasets are not large enough for diffusion model training such that data augmentation is often necessary. However, typical image data augmentation techniques, e.g., crop, rotate, flip, etc, may negatively affect the recognition (difficult to read) of character and word contents. In this work, we implement two simple strategies for document-level data augmentation, {\color{black} while mitigating the high-resolution challenges such as computational scalability issues in training diffusion models}, as shown in Fig.\ref{fig:DA}. 

The sub-window strategy divides the high-resolution images into several smaller domains, e.g., 1024x1024 images will be divided into 16 sub-images (256$\times$256) or 64 sub-images (128$\times$128). Using this way, we reduce the image resolution but upscale the dataset size fed to the diffusion models for better performance at a lower training cost. If the data is very sparse, we can consider the slide-window strategy, which is inspired by convolution operation in CNN, moving the sub-window with a specific stride. This strategy will significantly increase the amount of data which allows diffusion models to accurately capture the distribution of characters and words. For translation, we perform the same strategy for the source (noisy) data and obtain the corresponding target sub-images, and finally we ensemble all of them to obtain the whole cleaned images.  
\begin{algorithm}[h!]
\small
\caption{\hspace{-0.1cm} Diffusion model training in DECDM}
\begin{algorithmic}[1] \label{algo:1}
\STATE{\bf Requirement}: noise data from source domain, $\mathbf x^{(s)} \sim p_{s}(\mathbf x)$, clean data from target domain, $\mathbf x^{(t)} \sim p_{t}(\mathbf x)$.
\STATE Perform data augmentation for $\mathbf x^{(s)}$ and $\mathbf x^{(t)}$ 
\STATE Train source diffusion model $v_{\theta}^{(s)} (\mathbf x^{(s)}) \approx p_{s}(\mathbf x)$ and target diffusion model $v_{\theta}^{(t)} (\mathbf x^{(t)}) \approx p_{t}(\mathbf x)$ separately  
\STATE{\bf Return} trained source model $v_{\theta}^{(s)}$ and target model $v_{\theta}^{(t)}$
\end{algorithmic}
\end{algorithm}
\begin{algorithm}[h!]
\small
\caption{\hspace{-0.1cm} Unpaired image translation in DECDM}
\begin{algorithmic}[1] \label{algo:2}
\STATE{\bf Requirement}: data sample from source domain $\mathbf x^{(s)} \sim p_{s}(\mathbf x)$, source model $v_{\theta}^{(s)}$, target model, $v_{\theta}^{(t)}$, $t_0, t_1$
\STATE{\bf Encoding}: obtain latent embedding from source domain data via $\mathbf x^{(z)} = \mathcal{S}_{\textnormal{ODE}} (\mathbf x^{(s)}; v_{\theta}^{(s)}, t_0, t_1);$
\STATE{\bf Decoding}: obtain target domain data reconstructed from latent code via $\mathbf x^{(t)} = \mathcal{S}_{\textnormal{ODE}} (\mathbf x^{(z)}; v_{\theta}^{(t)}, t_1, t_0)$
\STATE{\bf Return}: $\mathbf x^{(t)}$
\end{algorithmic}
\end{algorithm}

\section{Experiments}
A set of experiments are provided to demonstrate the effectiveness of our DECDM. We first use a 2D synthetic example to show the cycle-consistent property and then demonstrate DECDM on various document enhancement tasks, including dirty document denoising and shadow removal. 
\subsection{2D Synthesis Examples}
\begin{figure}[h!]
    \centering
    \includegraphics[width=0.48\textwidth]{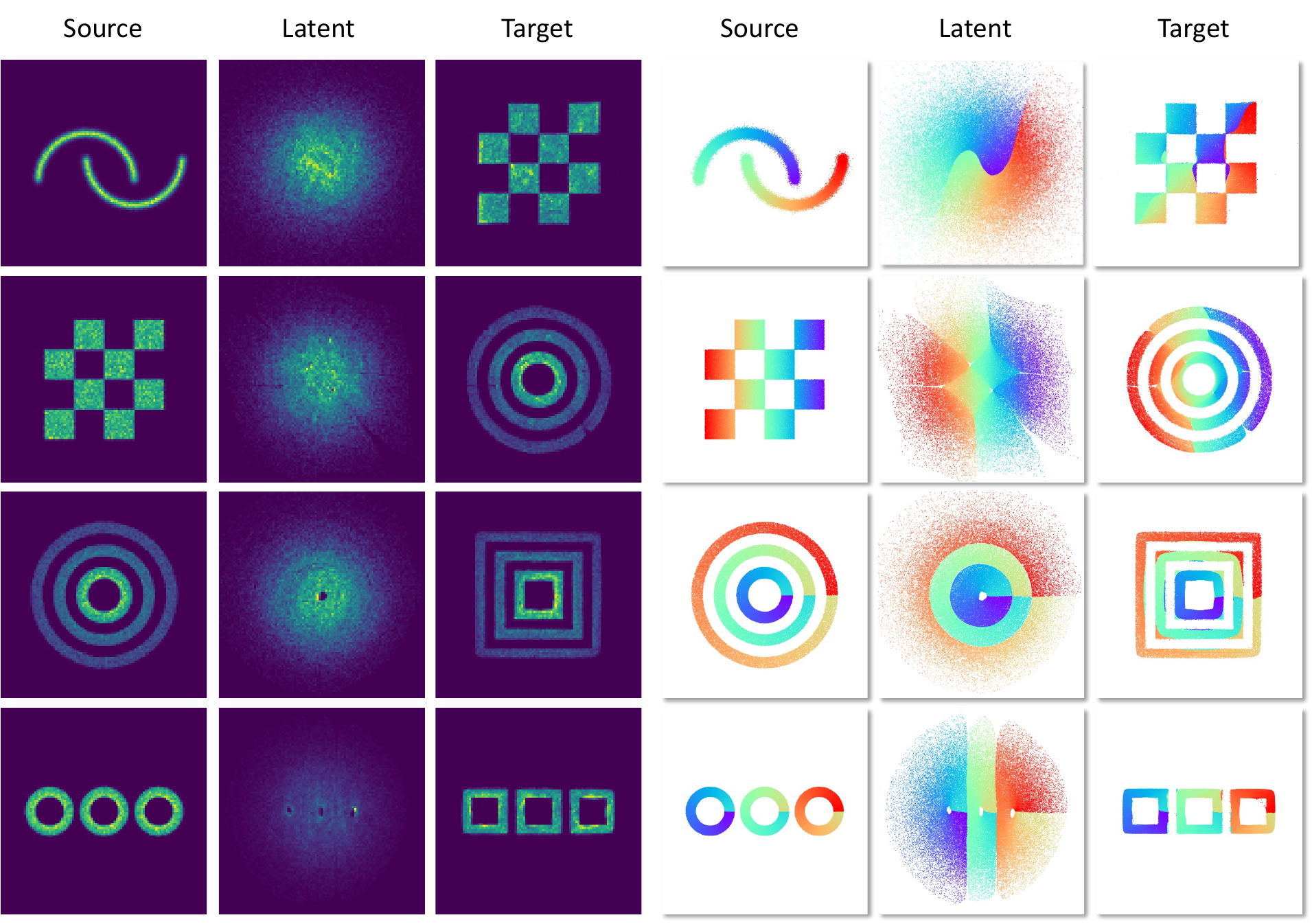}
    \caption{Distribution translation of synthetic datasets: from source datasets to latent representation via encoding, then from latent representation to target datasets via decoding.  (Left three) heatmap results and (Right three) scatter results with color configurations. }
    \label{fig:toy1}
\end{figure}

We perform domain distribution translation on two-dimensional synthetic datasets with complex shapes and configurations, as shown in Fig.~\ref{fig:toy1}. In this example, we use six 2D datasets (normalized to zero mean and identify covariance): Two Moons (TM); Checkerboards (CB); Concentric Rings (CR); Concentric Squares (CS); Parallel Rings (PR); and Parallel Squares (PS). The colors in Fig.~\ref{fig:toy1} are signed based on the point identities that can help check if a point in the source domain is blue, then its corresponding point in the target domain is also colored blue. To this end, we observed a smooth translation between the source and target domain with point identity preservation. For instance, on the second row in Fig.~\ref{fig:toy1}, the red points in the CR dataset are mapped to similar coordinates (relative location) in the target domain of the CS dataset. The latent space provides a disentangled representation of this domain translation. 
\vspace{-2mm}

\noindent {\bfseries Cycle Consistency Validation}. We demonstrate the cycle consistency using an example of domain translation from CR to PR, as shown in Fig.~\ref{fig:cycle}. We first train the cycle-consistent diffusion models for each domain (CR and PR) independently. Then starting from the CR dataset $\mathbf x^{(s)}$, we obtain the latent points $\mathbf x^{(z)}$ using reverse sampling and construct the target PR points $\mathbf x^{(t)}$ via direct sampling. The next step is the reverse direction, i.e., transforming the target PR points back to the latent and the source CR domain. Similarly, we transfer $\mathbf x^{(t)}$ to the latent points $\Tilde{\mathbf x}^{(z)}$ using reverse sampling and then reconstruct the source CR domain $\Tilde{\mathbf x}^{(s)}$ via direct sampling. After this multi-step trip, the source points are approximately mapped back to their original positions. From Fig.~\ref{fig:cycle}, we observed a similar color topology both in the latent and source domain. The reconstructed source points $\Tilde{\mathbf x}^{(s)}$ are highly consistent with the original source points $\mathbf x^{(s)}$. To further compare the difference, Table \ref{tab:cycle} shows quantitative evaluation results on cycle consistency among various cases. We use averaged L2 distance to measure the difference between the original points and the reconstructed points after cycle translation, e.g., "TM-CB" means TM $\rightarrow$ CB $\rightarrow$ TM. The results in Table \ref{tab:cycle} are negligibly small in terms of both the latent and source domains such that the cycle consistency is valid even without adding cycle-consistent loss \cite{zhu2017unpaired}.
    \vspace{-2mm}
\begin{table}[!h]
\footnotesize
\centering
\begin{tabular}{@{}ccccccc@{}}
\toprule
Distance & TM-CB  & CR-TB  & CR-CS & CR-PR   & PR-PS  & PS-CS  \\ \midrule
Latent  & 0.0128 & 0.0087 & 0.0101  & 0.0120 & 0.0092 & 0.0100   \\
Source  & 0.0122 & 0.0106 & 0.0082 & 0.0108 & 0.0143 & 0.0065 \\ \bottomrule
\end{tabular}
\vspace{-2mm}
\caption{Cycle consistency validation. Averaged L2 distance is used to measure the difference between original points and after-cycle translation on both latent and source domains. }
\vspace{-3mm}
\label{tab:cycle}
\end{table}
\vspace{-3mm}
\begin{figure}[h!]
    \centering
    \includegraphics[width=0.48\textwidth]{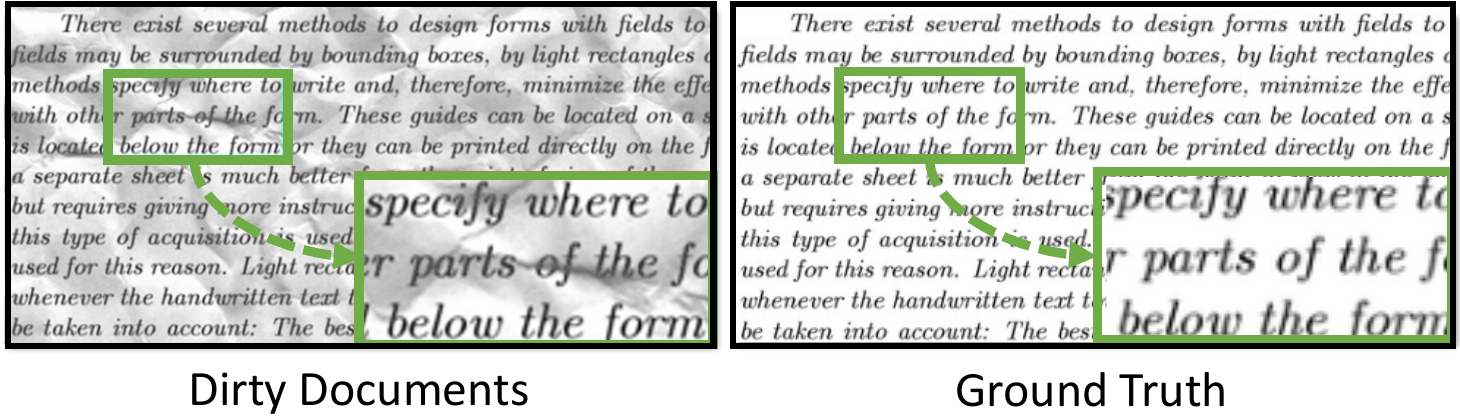}
    \vspace{-6mm}
    \caption{Visualization of DatasetA: (Left) raw document-level image and (Right) ground truth, which is the clean image. }
    \label{fig:datasetA}
\end{figure}
    \vspace{-6mm}
\begin{figure}[h!]
    \centering
    \includegraphics[width=0.48\textwidth]{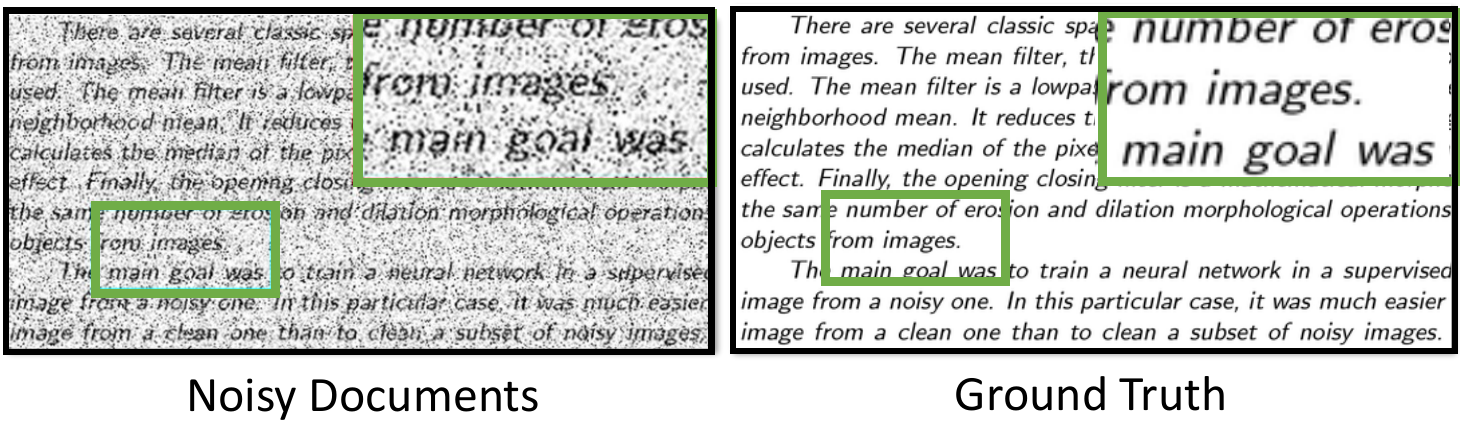}
    \vspace{-6mm}
    \caption{Visualization of DatasetB: (Left) noisy document-level image and (Right) ground truth. }
    \label{fig:datasetB}
\end{figure}
\begin{figure*}[h!]
    \centering
    \includegraphics[width=0.96\textwidth]{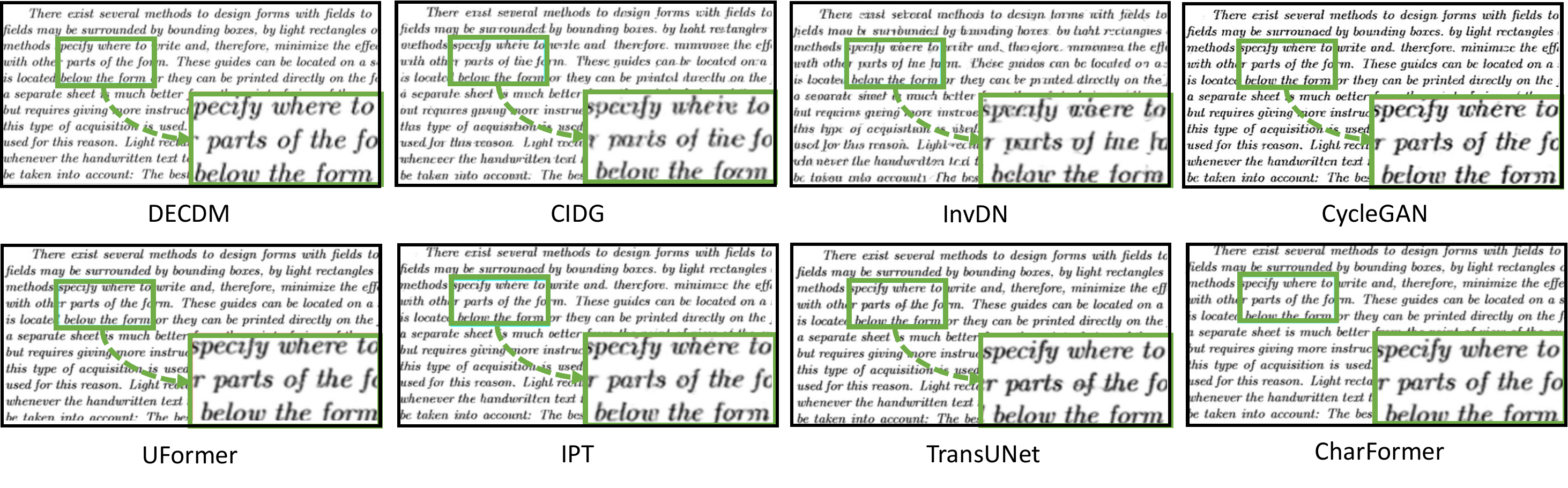}
    \vspace{-4mm}
    \caption{Qualitative evaluations and comparisons on {\em DatasetA} which is dirty document denoising.}
    \label{fig:dataset_a_results}
\end{figure*}
\begin{figure*}[h!]
    \centering
    \includegraphics[width=0.96\textwidth]{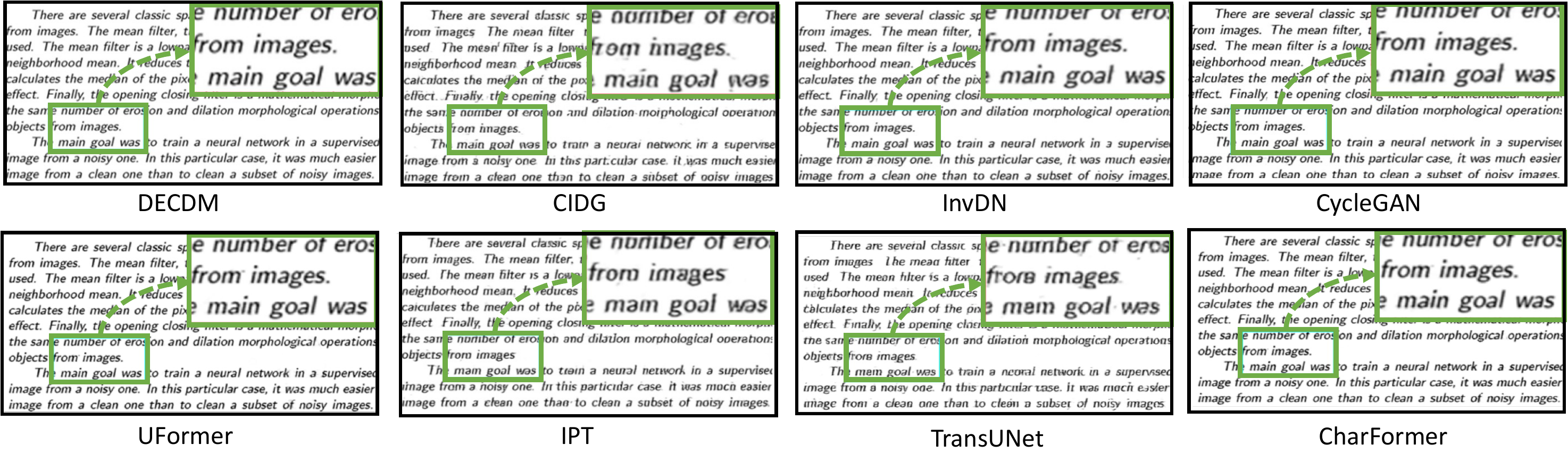}
    \vspace{-4mm}
    \caption{Qualitative evaluations and comparisons on {\em DatasetB} which is dirty document denoising.}
    \label{fig:dataset_b_results}
\end{figure*}

\subsection{Dirty Document Denoising}
\noindent {\bfseries Datasets}. In this case, we apply our DECDM for denoising dirty documents by leveraging the benchmark datasets \emph{denoising-dirty-document}\footnote{{https://www.kaggle.com/competitions/denoising-dirty-documents}}, which consists of printed English words in 18 different fonts. The original datasets include noisy raw document-level images with uneven backgrounds, e.g., watermarks, messy artifacts, etc. We name the original datasets as {\em DatasetA: Dirty Document}. {\color{black} There are 144 data for training and 72 data for testing in the original setting. We use this setting for evaluating all the methods.} To increase the complexity, we also create {\em DatasetB: Noisy Document} by adding speckle noise and Gaussian noise on the ground truth. The noise means $\mu$ is 0 and variance $\sigma$ is 5, which follows the setting in \cite{shi2022charformer}. Fig.~\ref{fig:datasetA} shows one of the raw document-level images and the corresponding clean image in DatasetA. Fig.~\ref{fig:datasetB} shows the noisy document-level image in DatasetB. 

\vspace{2mm}
\noindent {\bfseries Baselines}. We compare our DECDM with multiple competitive baseline methods, including GAN/CNN-based methods, CIDG \cite{zhang2020novel}, InvDN \cite{liu2021invertible}, CycleGAN\cite{sharma2018learning}, and some Transformer-based methods, i.e., UFormer \cite{wang2022uformer}, IPT \cite{chen2021pre}, TransUNet\cite{chen2021transunet} and CharFormer\cite{shi2022charformer}. Note that most of these state-of-the-art methods are proposed for general image denoising or restoration, not specifically designed for document denoising. Thus, we use the same training environment and datasets for all the methods and report the results if they have already been provided in their work \cite{shi2022charformer}. We perform a slide-window strategy for data augmentation in this case and all the experiments and comparisons are done on one NIVIDA Tesla V100 GPU. Appendix \ref{sec:samples} presents some synthesized samples drawn from the trained source (noisy datasets) and target models (clean datasets). 
\begin{table}[!h]
\scriptsize
\centering
\begin{tabular}{@{}c|ccc|ccc@{}}
\toprule
\multirow{2}{*}{Method} & \multicolumn{3}{c|}{DatasetA} & \multicolumn{3}{c}{DatasetB} \\ \cmidrule(l){2-7} 
                        & PSNR$\uparrow$    & SSIM$\uparrow$     & AC$\uparrow$    & PSNR$\uparrow$    & SSIM$\uparrow$      & AC$\uparrow$    \\ \midrule
Raw Data                & 16.33   & 0.7978   & 0.6931  & 13.03   & 0.2852   & -      \\
CIDG \cite{zhang2020novel}                    & 21.88   & 0.8871   & 0.7559  & 20.65   & 0.8623   & 0.2471  \\
InvDN \cite{liu2021invertible}                    & 22.40    & 0.8807   & 0.8374  & 20.49   & 0.8077   & 0.5917  \\
CycleGAN \cite{sharma2018learning}                & 23.66   & 0.8857   & 0.8319  & 20.97   & 0.8470    & 0.6409  \\
UFormer \cite{wang2022uformer}                 & 23.86   & 0.8970    & 0.8326  & 21.01   & 0.8221   & 0.6693  \\
IPT \cite{chen2021pre}                     & 23.72   & {\bf0.9027}   & 0.856   & 21.94   & 0.8293   & 0.6854  \\
TransUNet \cite{chen2021transunet}               & 23.92   & 0.8998   & {\bf 0.8621}  & 20.83   & 0.8592   & 0.5579  \\
CharFormer \cite{shi2022charformer}              & {\bf24.08}   & 0.8985   & 0.8553  & {\bf 21.07}   & {\bf 0.8637}   & {\bf 0.7259}  \\ \midrule
DECDM   & {\bf 24.30}       & {\bf 0.9058}        & {\bf 0.8714}       & {\bf 21.12}       & {\bf 0.8631}        & {\bf 0.7438}       \\ 
\bottomrule
\end{tabular}
\vspace{-2mm}
\caption{Quantitative evaluation results on average PSNR, SSIM and OCR accuracy (AC). The best two results are highlighted in bold black.} 
\vspace{-3mm}
\label{tab:case1}
\end{table}

\noindent {\bfseries Metrics}. We introduce two commonly used metrics to evaluate the document-level denoising performance, i.e., peak signal-to-noise ratio (PSNR) and the structural similarity index measure (SSIM). Note that ``$\uparrow$'' represents the higher the metric the higher image quality. Additionally, we introduce a metric for evaluating the character-level quality, i.e., optical character recognition (OCR) accuracy (AC). This metric allows us to validate if the denoising algorithms improve the OCR\footnote{The public OCR tools can be accessed via {https://www.ocr2edit.com}} performance compared to dirty documents. 

\begin{figure*}[h!]
    \centering
    \includegraphics[width=0.95\textwidth]{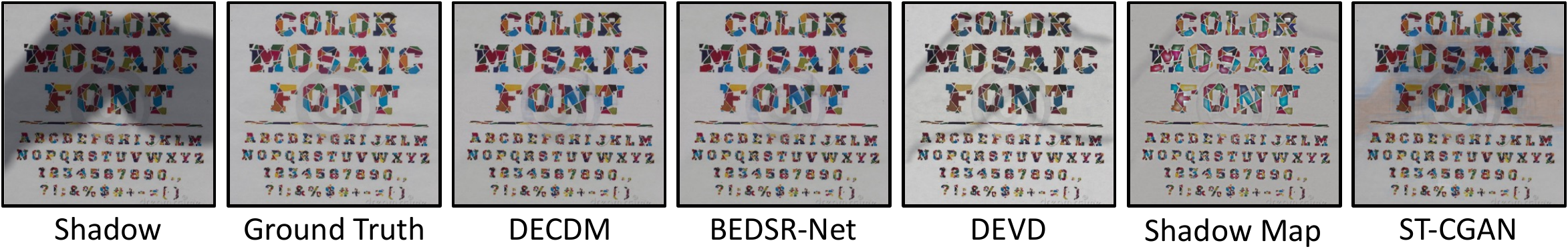}
    \vspace{-2mm}
    \caption{Qualitative evaluation and visual comparison of competing baseline methods on document {\color{black}shadow} removal task.}
    \label{fig:shallow1}
\end{figure*}
 
\begin{table*}[!h]
\footnotesize
\centering
\begin{tabular}{@{}c|cc|cc|cc|cc|cc@{}}
\toprule
Method        & \multicolumn{2}{c|}{SDSRD \cite{lin2020bedsr}} & \multicolumn{2}{c|}{RDSRD \cite{lin2020bedsr}} & \multicolumn{2}{c|}{
SM Datasets \cite{bako2016removing}} & \multicolumn{2}{c|}{DVED Datasets \cite{kligler2018document}} & \multicolumn{2}{c}{WF Datasets \cite{jung2018water}} \\ \midrule
              & PSNR $\uparrow$        & SSIM $\uparrow$         & PSNR $\uparrow$         & SSIM $\uparrow$        & PSNR $\uparrow$           & SSIM $\uparrow$           & PSNR $\uparrow$       & SSIM $\uparrow$       & PSNR $\uparrow$           & SSIM $\uparrow$            \\ \cmidrule(l){2-11} 
Raw Shadow Images      & 22.80        & 0.8992      & 21.73       & 0.8093      & 28.45          & 0.9742         & 19.31      & 0.8429      & 20.35           & 0.8850           \\
{\color{black}Shadow} Map \cite{bako2016removing}  & 31.55       & 0.9658      & 28.24       & 0.8664      & {\bf 35.22}          & {\bf 0.9823}         & 29.66      & 0.9051      & 23.70            & 0.9015          \\
DVED \cite{kligler2018document}          & 22.03       & 0.8435      & 22.53       & 0.7056      & 26.50           & 0.8381         & 26.45      & 0.8481      & 24.45           & 0.8332          \\
Water Filling \cite{jung2018water} & 17.06       & 0.8226      & 14.45       & 0.7054      & 13.88          & 0.8059         & 19.21      & 0.8724      & {\bf 28.49}           & 0.9108          \\
ST-CGAN \cite{wang2018stacked}        & 39.38       & 0.9834      & 30.31       & 0.9016      & 29.12          & 0.9600           & 25.92      & 0.9062      & 23.71           & 0.9046          \\
BEDSR-Net \cite{lin2020bedsr}     & {\bf 43.59}       & {\bf 0.9935}      & {\bf 33.48}       & {\bf 0.9084}      & {\bf 35.07}          & {\bf 0.9809}         & {\bf 32.90}       & {\bf 0.9354}      & {27.23}           & {\bf 0.9115}          \\ \midrule
DECDM         & {\bf 45.73}       & {\bf 0.9932}      & {\bf 37.21}       & {\bf 0.9143}      & {34.95}          & 0.9642         & {\bf 35.01}      & {\bf 0.9521}      & {\bf 29.87}           & {\bf 0.9112}         \\ \bottomrule
\end{tabular}
\vspace{-2mm}
\caption{Quantitative evaluation results on PSNR and SSIM. We compare our DECDM with BEDSR-Net \cite{lin2020bedsr}, ST-CGAN \cite{wang2018stacked}, Water Filling \cite{jung2018water}, DVED \cite{kligler2018document}, and {\color{black} Shadow} Map \cite{bako2016removing} methods. The best two results are highlighted in black bold.} 
\vspace{-3mm}
\label{tab:shadow}
\end{table*}

\vspace{2mm}
\noindent {\bfseries Qualitative Evaluation}. We first visualize the denoising results by using DECDM and compare it with other baseline methods. Fig.~\ref{fig:dataset_a_results} and Fig.~\ref{fig:dataset_b_results} show the qualitative performance on {\em DatasetA} and {\em DatasetB} respectively. DECDM can effectively remove messy dirties and even backgrounds and perform high-quality document-level image denoising. Unlike some methods, e.g., CycleGAN, InvDN, and TransUNet with character-level damages, DECDM well recognizes the character style and topology, which can be clearly seen in the zoom-in sub-figures in Fig.~\ref{fig:dataset_a_results}. As an unpaired method, DECDM shows competitive performance compared to the transformer-based methods, e.g., CharFormer and UFormer, which strongly rely on paired supervision. {\color{black} More ablation studies are provided in Appendix \ref{sec:additional_exp}}.

\vspace{2mm}
\noindent {\bfseries Quantitative Evaluation}.
Table \ref{tab:case1} shows the quantitative comparisons between DECDM and state-of-the-art baseline methods on both datasets. Clearly, DECDM shows outperformed results, specifically the AC metric, in both datasets. Compared with GAN/INN models, transformer-based models perform competitively, e.g., CharFormer in DatasetB but it will fail in the unsupervised setting.

\subsection{Document Shadow Removal}
\noindent {\bfseries Datasets}.  Although there exist a few datasets for document image shadow removal, they are only used for evaluation on a small scale. In this example, we consider the following five datasets ranging from small-scale to large-scale such that we can provide a comprehensive validation. 
\begin{itemize}[leftmargin=2ex, nosep, noitemsep]
    \item SDSRD datasets \cite{clausner2017icdar2017, lin2020bedsr}: 8309 paired images from 970 documents, including synthetic, diverse contexts and lighting. 7533 for training and 776 for testing. 
    \item RDSRD datasets \cite{lin2020bedsr}: 540 paired images of 25 documents, including newspaper, slides, and paper, under different lighting conditions.
    \item {\color{black}Shadow} Map (SM) datasets \cite{bako2016removing}: 81 paired images with light shadows/text only. 
    \item DEVD datasets \cite{kligler2018document}: 300 paired document-level images, including dark shadows and colorful symbols. 
    \item Water-Filling datasets \cite{jung2018water}: 87 high-quality paired images including multi-cast shadows. 
\end{itemize}
\vspace{2mm}
\noindent {\bfseries Baselines}. We compared our DECDM with five state-of-the-art methods, including BEDSR-Net \cite{lin2020bedsr}, ST-CGAN \cite{wang2018stacked}, Water Filling \cite{jung2018water}, DVED \cite{kligler2018document}, and {\color{black} Shadow} Map \cite{bako2016removing} methods. For a fair comparison, we used the publicly available source codes or reported results provided by the authors. We evaluate the compared methods from visual quality using the PSNR and SSIM metrics, as suggested by \cite{lin2020bedsr}.  
\vspace{2mm}

\noindent {\bfseries Qualitative and Quantitative Evaluation}. For visual comparison, Fig. \ref{fig:shallow1} shows several shadow removal results of the compared methods. DEVD \cite{kligler2018document} and ST-CGAN \cite{wang2018stacked} exhibit remaining shadow edges and {\color{black} Shadow} Map \cite{bako2016removing} performs better than those two but still shows the shadow. DECDM close to BEDSR-Net \cite{lin2020bedsr} shows ideal performance without seeing shadow edges. Quantitatively, DECDM outperforms other baselines on most datasets as shown in Table \ref{tab:shadow}.  For SM datasets, {\color{black} Shadow} Map performs best but its result is worse than the other baselines in the other four datasets. BEDSR-Net is a competitive method that achieves promising results but it strongly relies on the pair datasets. On the contrary, DECDM is more flexible and robust without the assumption of pair knowledge such that we can easily deploy it in more real-world scenarios. {\color{black} We also provide a detailed analysis of the effect of data augmentation strategies in Appendix \ref{sec:additional_exp}.}

\section{Related Work}
\paragraph{Document Enhancement.} Deep learning has enabled many approaches for enhancing the quality of document-level images \cite{anvari2021survey}. Recent state-of-the-art methods in document enhancement are summarized in Table \ref{tab:summary}, categorized by their supervision mechanism (paired or unpaired), backbone models (CNNs \cite{zhao2018skip, calvo2019selectional, li2021adaptive}, GANs \cite{sharma2018learning, gangeh2021end, lin2020bedsr, souibgui2020gan, xu2017learning}, and Transformers \cite{shi2022charformer, souibgui2022docentr}), and enhancement tasks (denoising, shadow removal, binarization, watermark removal, deblur, and defade). Although most methods perform well in one or multiple tasks, no single model can handle all types. Additionally, paired supervision is required, which is rarely met in real settings. While Cycle-GAN \cite{gangeh2021end,sharma2018learning} methods can mitigate this limitation, they still need to optimize for cycle consistency over two domains, leading to instability issues and potential data privacy leakage. Our proposed DECDM addresses these challenges by enabling unpaired translation, cycle consistency, and data privacy protection.

\vspace{-5mm}
\paragraph{Diffusion Models.} Diffusion models are a family of generative models that have gained much attention recently due to their superior performance in text-guided image synthesis \cite{ruiz2022dreambooth, balaji2022ediffi, feng2022training}, e.g., Stable Diffusion \cite{rombach2022high}, DALL·E 2 \cite{ramesh2022hierarchical}, and Imagen \cite{saharia2022photorealistic}. These works are built upon the foundation of diffusion models, including score-based methods \cite{song2019generative, song2020score} that match with Langevin dynamics, denoising diffusion probabilistic models (DDPMs) \cite{ho2020denoising, sohl2015deep} that parameterize the ELBO objective with Gaussian, and denoising diffusion implicit models (DDIMs) \cite{song2020denoising} that accelerate DDPM inference via non-Markovian processes. Recent works have leveraged diffusion models for image editing \cite{choi2021ilvr, kwon2022diffusion, wu2022unifying, sasaki2021unit}, composition \cite{meng2021sdedit, zhao2022egsde}, and restoration tasks \cite{kawar2022denoising, saharia2022photorealistic} with promising performance. However, these methods mostly relied on joint training by leveraging both datasets directly.  Our DECDM performs a decoupled mechanism by applying separate, pretrained diffusion models and leveraging the geometry of the shared space for document image translation. To the best of our knowledge, DECDM is the first work to apply diffusion models for document enhancement via unpaired image translation, inspired by these studies.

\section{Conclusions}
\label{sec:conclusions} 
DECDM provides an unsupervised end-to-end solution for document image enhancement that offers several advantages over existing state-of-the-art methods, including adaptability to new domain pairs and data privacy protection. These unique capabilities make DECDM a more robust, safe, and scalable solution for improving OCR performance in a wide range of document enhancement tasks. Future works aim to address the current limitations caused by data sparsity, augmentation, and character/word context recognition. We will also integrate OCR into the training pipeline to pursue better character and word recognition. 

{\small
\bibliographystyle{ieee_fullname}
\bibliography{egbib}
}

\clearpage 
\appendix 
\section{Details of DDIM ODE Solver}
\label{sec:ode_solver}
\subsection{Diffusion Models} Diffusion Denoising Probabilistic Models (DDPM) \cite{sohl2015deep,ho2020denoising} aim at modeling a distribution $p_{\theta}(\mathbf x_0)$ to approximate the data distribution $q(\mathbf x_0)$. The forward process performs a progressing procedure from $\mathbf x_0$ to $\mathbf x_T$ via a Markov chain, where we generate the latent variables $\mathbf x_1, ...,\mathbf x_T$ by gradually adding noise to the data via Gaussian transition. When $T$ is large enough, the last noise vector $\mathbf x_T$ nearly follows an isotropic Gaussian distribution. 

The forward process has a simple closed-form solution that expresses the latent variable $\mathbf x_t, t \in \{0,..., T\}$ as a linear combination of noise and $\mathbf x_0$ \cite{ho2020denoising}:
\begin{equation}
    \mathbf x_t = \sqrt{\alpha_t}\mathbf x_0 + \sqrt{1-\alpha_t}
\epsilon_t, \quad \epsilon_t \sim \mathcal{N}(0, \mathbf I),
\end{equation}
where $\alpha_t$ is referred to as the noising schedule which defines the amount of noise present at each intermediate timestep, $0 = \alpha_T < \alpha_{T-1}<...,<\alpha_1 < \alpha_0 = 1$. Each refinement step consists of an application of a neural network $f_{\theta}(\mathbf x, t)$ on the current sample $\mathbf x_t$, followed by a random Gaussian noise perturbation, obtaining $\mathbf x_{t-1}$. The network is trained for a simple denoising objective, aiming for $f_{\theta}(\mathbf x_t, t) = \epsilon_{\theta}^{(t)}(\mathbf x_t) \approx \epsilon_t$.

Sampling from distribution $q(\mathbf x_0)$ is defined by a reverse process, from isotropic Gaussian noise $\mathbf x_T$ to data, which is refined iteratively through $t \le T$ passes through the network. There are various sampling strategies \cite{song2020denoising,nichol2021improved} that define the process of merging the noise prediction $\epsilon_{\theta}^{(t)}(\mathbf x_t)$ and current sample $\mathbf x_t$ to produce the previous sample $\mathbf x_{t-1}$. The final $\mathbf x_0$ sample is the resultant generated image. 

\subsection{DDIM Inversion}
Unlike the commonly used DDPM, the generative sampling process in DDIMs is defined in a non-Markovian manner, 
\begin{equation}
    \mathbf x_{t-1} = \sqrt{\frac{\alpha_{t-1}}{\alpha_t}} \mathbf x_t + \left( \sqrt{\frac{1-\alpha_{t-1}}{\alpha_{t-1}}} - \sqrt{\frac{1-\alpha_t}{\alpha_t}}\right) \epsilon_{\theta}^{(t)}(\mathbf x_t)  \label{eq:ddim_1}
\end{equation}
which can be used for inversion, based on the assumption that the ordinary differential equation (ODE) process can be reversed in small steps:
\begin{equation}
    \mathbf x_{t+1} = \sqrt{\frac{\alpha_{t+1}}{\alpha_t}} \mathbf x_t + \left( \sqrt{\frac{1-\alpha_{t+1}}{\alpha_{t+1}}} - \sqrt{\frac{1-\alpha_t}{\alpha_t}}\right) \epsilon_{\theta}^{(t)}(\mathbf x_t).  \label{eq:ddim_2}
\end{equation}
Thus, the diffusion process is performed in the reverse direction, deterministically noising an image to obtain the initial noise vector. In other words, DDIM inversion achieves $\mathbf x_0 \rightarrow \mathbf x_T$ instead of $\mathbf x_T \rightarrow \mathbf x_0$.  

Empirically, the error of DDIM inversion is reasonably small since Eq.~\ref{eq:ddim_2} can be treated as an Euler method over the following ODE, which is up to discretization errors of the ODE solvers:
\begin{equation}
    \text{d}\hat{\mathbf x}(t) = \epsilon_{\theta}^{(t)} \left(\frac{\hat{\mathbf x}(t)}{\sqrt{\sigma^2 + 1}} \right) \textup{d} \sigma(t) \label{eq:ode}
\end{equation}
where $\hat{\mathbf{x}} = \mathbf x /\sqrt{\alpha}$ and $\sigma = \sqrt{1-\alpha}/\sqrt{\alpha}$. However, in practice, a slight error is incorporated in every step, and eventually, the accumulated error might be non-negligible.  In some cases, the obtained noise vector might be out of the Gaussian assumption. Importantly, the ODE in Eq.~\eqref{eq:ode} with the optimal model $\epsilon_{\theta}^{(t)}$ has an equivalent probability flow ODE corresponding to the variance exploding SDE \cite{song2020score}. Although the ODE solver has a reasonably small error, we can leverage recent developments in higher-order ODE solvers, such as the DPM-solver \cite{lu2022dpm}, the Exponential Integrator \cite{zhang2022fast}, and the second-order Heun Solver \cite{karras2022elucidating} that generalize DDIMs can also be used in our case. 

\subsection{DDIM Cycle-Consistency}

DDIMs invent a particular parameterization of the diffusion process, that creates a smooth, deterministic, and {\em reversible} mapping between images and their latent representations. This mapping is captured using the solution to a so-called probability flow (PF) \cite{song2020score}, ordinary differential equation (ODE). Translation with DECDM on a source-target pair requires two different PF ODEs: the source PF ODE converts input images to the latent space; while the target ODE then synthesizes images in the target domain. As PF ODEs are used, the cycle consistency property is guaranteed as validated in Proposition 4. In practice, even with discretization error, DECDM incurs almost negligible cycle inconsistency, as shown in our empirical experiments on 2D synthesis examples (see Table 2).

\section{Details of Training Objectives}
\label{sec:training}
The DECDM training is equivalent to training a score-based model \cite{song2019generative,song2020score}. Given samples from a data distribution $q(\mathbf x_0)$, diffusion models attempt to learn a model distribution $p_{\theta}(\mathbf x_0)$ that approximates $q(\mathbf x_0)$ and is easy to sample from. Specifically, diffusion models are latent variable models of the form:
\begin{equation}
    p_{\theta}(\mathbf x_0) = \int p_{\theta}(\mathbf x_{0:T}) \textup{d} \mathbf x_{1:T}
\end{equation}
where 
\begin{equation}
    p_{\theta}(\mathbf x_{0:T}) = p_{\theta}(\mathbf x_T)\prod_{t=1}^Tp_{\theta}^{(t)}(\mathbf x_{t-1}|\mathbf x_t)
\end{equation}
where $\mathbf x_1,...,\mathbf x_T$ are latent variables in the same sample space as $\mathbf \mathbf x_0$. The parameters $\theta$ are trained to approximate the data distribution $q(\mathbf x_0)$ by maximizing a variational lower bound:
\begin{equation}
    \begin{aligned}
    \max_{\theta} \mathbb E_{q(\mathbf x_0)} [\log p_{\theta}(\mathbf x_0)] &\le \max_{\theta}\mathbb{E}_{q(\mathbf x_0,...,\mathbf x_T)}[\log p_{\theta}(\mathbf x_{0:T}) 
    \\ & -\log q(\mathbf x_{1:T} | \mathbf x_0) ]
    \end{aligned}
\end{equation}
where $q(\mathbf x_{1:T}| \mathbf x_0)$ is some inference distribution over the latent variables. The training objective can be reformulated when the conditional distributions are modeled as Gaussian with trained mean functions and fixed variances:
\begin{equation}
    \begin{aligned}
    \mathcal{L}(\epsilon_{\theta}) &= \sum_{t=1}^T \mathbb{E}_{\mathbf x_0 \sim q(\mathbf x_0), \epsilon_t \sim \mathcal{N}(0, I)} [ \| \epsilon_{\theta}^{(t)}  (\sqrt{\alpha_t} \mathbf x_0  \\ &+\sqrt{1-\alpha_t} \epsilon_t ) - \epsilon_t \|_2^2 ]
    \end{aligned}
\end{equation}
From the above formulation, we know that the resulting noise prediction function $\epsilon_{\theta}^{(t)}$, are equivalent to the score networks $\mathbf s_{t, \theta}$ in \cite{ho2020denoising, song2020score}.

\section{Synthesized Samples}
\label{sec:samples}
\begin{figure}[h!]
    \centering
    \includegraphics[width=0.49\textwidth]{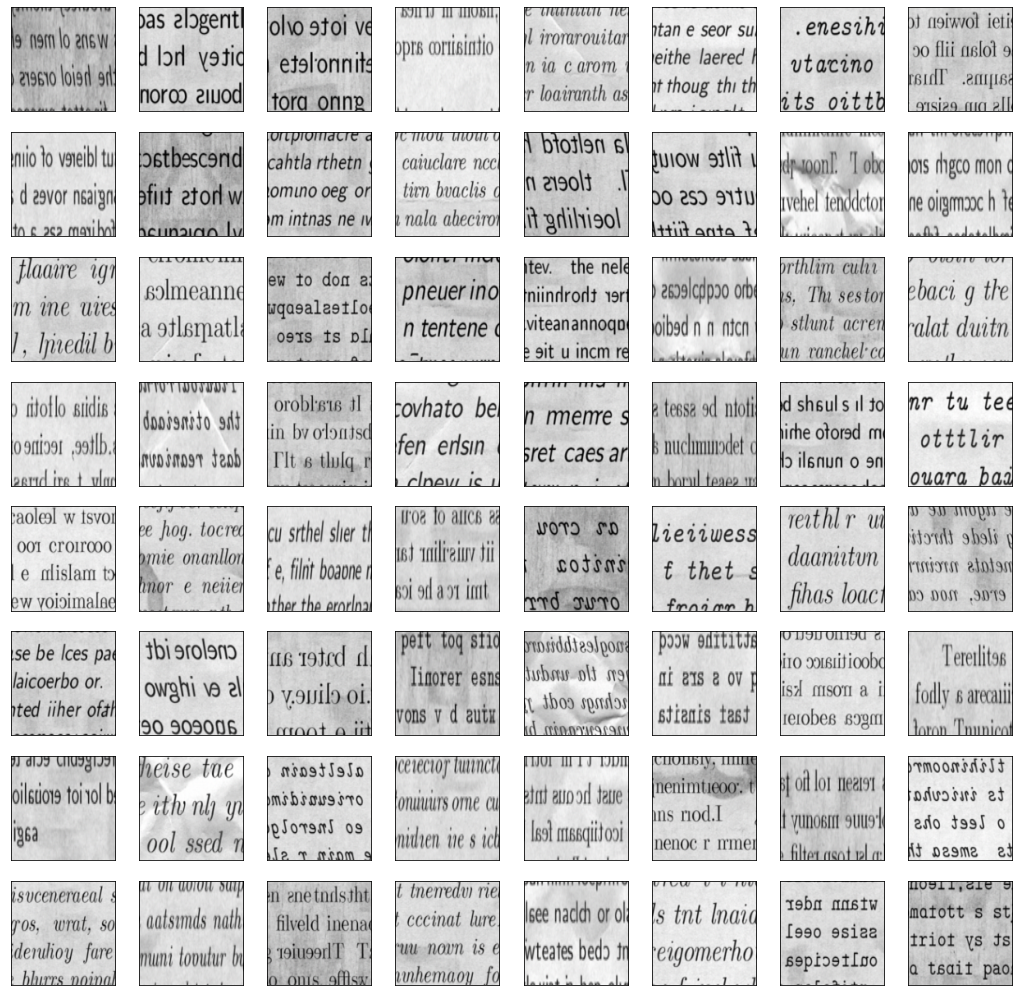}
    \vspace{-4mm}
    \caption{Synthesized noisy samples from dirty document datasets}
    \label{fig:noisy}
\end{figure}

\begin{figure}[h!]
    \centering
    \includegraphics[width=0.49\textwidth]{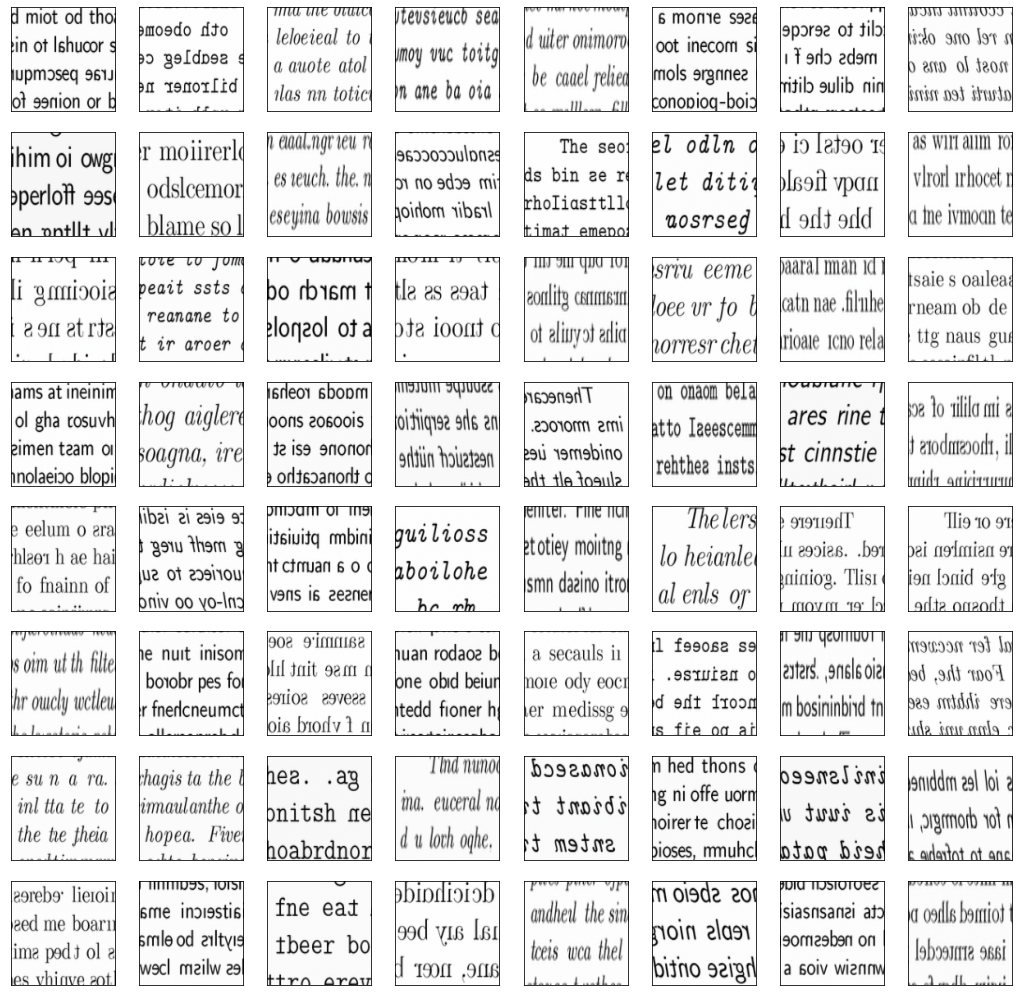}
    \vspace{-4mm}
    \caption{Synthesized clean samples from dirty document datasets}
    \label{fig:clean}
\end{figure}

DECDM consists of two trained models: the source model and the target model. Both models can also be used for synthesized data generation. Since both models are trained independently we can use them to generate noisy samples and clean samples separately, as shown in Fig.\ref{fig:noisy} and \ref{fig:clean}. In these two cases, the data augmentation strategy allows the rotation and flip such that the generated character is not shown as normal. We can clearly see the character feature without obvious damage and also capture some specific font styles.  The generated noisy samples successfully capture the noise distributions, which may be used for building noisy-clean pairs.  On the contrary, the clean samples drawn from the target models show clear character without any degradation.  

\begin{table*}[!h]
\small
\centering
\begin{tabular}{@{}c|ccc|ccc@{}}
\toprule
\multirow{2}{*}{Method} & \multicolumn{3}{c|}{DatasetA} & \multicolumn{3}{c}{DatasetB} \\ \cmidrule(l){2-7} 
                        & PSNR$\uparrow$    & SSIM$\uparrow$     & AC$\uparrow$    & PSNR$\uparrow$    & SSIM$\uparrow$      & AC$\uparrow$    \\ \midrule
DECDM without data augmentation   & {23.71}       & {0.8902}        & {0.8501}       & {20.49}       & {0.8390}        & {0.6876}       \\ 
DECDM with slide-window (64x64)  & {23.95}       & {0.8953}        & {0.8692}       & {20.87}       & {0.8481}        & {0.6943}  \\ 
DECDM with slide-window (128x128)    & {24.30}       & {0.9058}        & {0.8714}       & {21.12}       & {0.8631}        & {0.7438}       \\ 
DECDM with slide-window (256x256)    & {24.11}       & {0.9027}        & {0.8689}       & {20.96}       & {0.8607}        & {0.7302}       \\ 
\bottomrule
\end{tabular}
\caption{Effect of slide-window strategy with DECDM on dirty document denoising datasets.} 
\label{tab:ablation1}
\end{table*}

\section{Discussion of Data Privacy}
\label{sec:privacy}
Our method solves the data privacy issue by unpaired individual training and transferring encoded latent variables between two parties. Since both source and target diffusion models are trained independently, noise or clean images are only visible to the individual user. During translation, only latent variables are shared instead of noisy images, thus both noise and clean images are private to the two parties. For example, user A is the owner of the source domain and user B is the owner of the target domain. User A intends to translate the source images to the target domain in a private manner without releasing the source dataset. User B also wishes to make the target dataset private. In such a case, user  Acan can simply train a diffusion model with the source data, encode the data to the latent space, and only transmit the latent codes to user B. Then user B can use the pretrained diffusion models (using the target data) to convert the received latent code to a target image and send back to user A. The process only requires shared latent code from user A and a pretrained model from user B, which can be finished in a private platform, and both source and target datasets are private to the two parties. This is a significant advantage of DECDM over alternate methods, as we enable strong privacy protection of the datasets.

\section{More Details about Experiments}
\label{sec:additional_exp}

\subsection{Ablation Study on Denoising}
Table \ref{tab:ablation1} shows the ablation experiments to explore the effect of data augmentation strategy, such as the sub-window method on dirty document denoising datasets. In this case, we consider four cases: original DECEM without data augmentation, DECEM with slide-window with various window sizes ranging from 64x64, 128x128 to 256x256. As the window size increases, the total amount of augmented data decreases. The results show that the sliding window improves the performance of the original DECDM in terms of all evaluation metrics on both datasets. Smaller window size with larger datasets does not show the best performance. In contrast,  a moderate window size (128x128) outperforms the other two options in these cases.

\begin{table}[!h]
\small
\centering
\begin{tabular}{@{}c|cc@{}}
\toprule
\multirow{2}{*}{Method}        & \multicolumn{2}{c}{SDSRD \cite{lin2020bedsr}} \\ \cmidrule(l){2-3}
              & PSNR $\uparrow$        & SSIM $\uparrow$         \\ \midrule
DECDM without data augmentation        & {43.32}       & {0.9906}              \\ 
DECDM with sub-window (64x64)        & {43.51}       & {0.9912}              
\\
DECDM with sub-window (128x128)        & {45.73}       & {0.9932}
\\
DECDM with sub-window (256x256)        & {44.05}       & {0.9917} \\
\bottomrule
\end{tabular}
\caption{Effect of sub-window data augmentation strategy on document shadow removal.} 
\label{tab:ablation2}
\end{table}
\subsection{Ablation Study on Shadow Removal}
We also evaluate the performance of the data augmentation strategy on document shadow removal tasks. In this case, we choose the SDSRD \cite{lin2020bedsr} datasets for demonstration since there are relatively large dataset sizes (8309 paired images,  7533 for training, and 776 for testing). As mentioned in Section 3.4, we may prefer to use the sub-window strategy if the original data is not very sparse. Compared to the slide-window strategy, the sub-window strategy is easier and more efficient without additional ensembling steps. As shown in Table \ref{tab:ablation2}, we provide a comparison of various window sizes, such as 64x64, 128x128, and 256x256 to study the effect of sub-window strategy on shadow removal performance. Note that, DECDM with the 128x128 window size shows superior performance compared with the other two sizes, but the original DECDM also performs competitively since the original data size is already large enough to learn the potential source and target distribution. The improvement resulting from data strategy is not as significant as the case in dirty document denoising where the dataset size is small. Hence the slide-window strategy is recommended when the original dataset is sparse. 

\end{document}